\documentclass[runningheads, envcountsame, a4paper]{llncs}

\usepackage[pdftex]{graphicx}
\usepackage{subcaption}
\usepackage{mathtools}
\usepackage{blindtext}
\usepackage{soul}
\usepackage{xcolor}
\usepackage[hyphens]{url}
\usepackage{amsmath,amsfonts,amsthm,bm} 
\usepackage{booktabs}
\usepackage{multirow}
\usepackage{enumerate}
\usepackage[linesnumbered,ruled]{algorithm2e}
\usepackage{pifont}
\usepackage[acronym]{glossaries}
\usepackage{wrapfig}
\usepackage[misc]{ifsym}
\usepackage{filecontents}

\newcommand{\comment}[1]{ }

\newcommand{\andrea}[1]{{\color{red}\textbf{Andrea}: #1}}

\newcommand{\cmark}{\ding{51}}
\newcommand{\xmark}{\ding{55}}
\newcommand{\m}[1]{\ensuremath{\mathrm{#1}}}

\newcommand\fig[1]{Fig.~\ref{#1}}

\SetCommentSty{mycommfont}
\SetKwInput{KwInput}{Require}                
\SetKwInput{KwOutput}{Output}              

\graphicspath{ {./img/} }

\newacronym{al}{AL}{Active Learning}
\newacronym{ml}{ML}{Machine Learning}
\newacronym{sota}{SOTA}{State-Of-The-Art}
\newacronym{pr}{PR}{PRopagate}

\begin{document}

\title{Stream-based Active Learning with Verification Latency in Non-stationary Environments}
\titlerunning{Active Learning with Verification Latency in Non-stationary Environments}

\tocauthor{Andrea~Castellani, Sebastian~Schmitt, Barbara~Hammer}

\newcommand{\orcidauthorA}{\orcidID{0000-0003-0476-5978}} 
\newcommand{\orcidauthorB}{\orcidID{0000-0001-7130-5483}} 
\newcommand{\orcidauthorC}{\orcidID{0000-0002-0935-5591}} 

\author{Andrea~Castellani (\Letter) \inst{1} \orcidauthorA  \and
Sebastian~Schmitt \inst{2} \orcidauthorB \and
Barbara~Hammer \inst{1} \orcidauthorC} 
\authorrunning{A. Castellani and S. Schmitt and B. Hammer}
%
\institute{Bielefeld University, \\
\email{\{acastellani,bhammer\}@techfak.uni-bielefeld.de}\\ 
\and
Honda Research Institute Europe, \\
\email{sebastian.schmitt@honda-ri.de}
}
\maketitle              
\begin{abstract} 
Data stream classification is an important problem in the field of machine learning.
Due to the non-stationary nature of the data where the underlying distribution changes over time (\textit{concept drift}), the model needs to continuously adapt to new data statistics.
Stream-based Active Learning (AL) approaches address this problem by interactively querying a human expert to provide new data labels for the most recent samples, within a limited budget.
Existing AL strategies assume that labels are immediately available, while in a real-world scenario the expert requires time to provide a queried label (\textit{verification latency}), and
by the time the requested labels arrive they may not be relevant anymore. 
In this article, we investigate the influence of finite, time-variable, and unknown verification delay, in the presence of concept drift 
on AL approaches.
We propose \textit{PRopagate (PR)}, a latency independent utility estimator which also predicts the requested, but not yet known, labels.
Furthermore, we propose a drift-dependent dynamic budget strategy, which uses a variable distribution of the labelling budget over time, after a detected drift.
Thorough experimental evaluation, with both synthetic and real-world non-stationary datasets, and different settings of verification latency and budget are conducted and analyzed. We empirically show that the proposed method consistently outperforms the state-of-the-art.
Additionally, we demonstrate that with variable budget allocation in time, it is possible to boost the performance of AL strategies, without increasing the overall labeling budget.

\end{abstract}
%
%
%
\section{Introduction}
The growing digitization of industrial processes leads to an ever-increasing data stream recorded by many sensors.
In potentially critical settings, there is an interest to continuously monitor the data stream and  analyze data samples at the time they are recorded e.g., in order to detect failing machines or determine the current operating state of a machine in a factory setting. Data-driven online machine learning techniques can address such tasks, but 
require labeled data samples for task-specific training \cite{fahy2022scarcity}.
Realistic data streams are often non-stationary and  the underlying data statistics might change over time, so-called \textit{concept drifts} \cite{Gama2014ASO}. This can render labels and trained models outdated (in particular in the case of so-called real concept drift) and leads to the need to manually relabel data samples and retrain models continuously. 

Stream-based \acrfull{al} approaches address this problem by interactively querying an oracle e.g., a human expert, to provide new data labels for particularly informative training samples which arrive over time.
Since acquiring new labels is costly, models  typically operate based on an only limited total budget for new labels.
Therefore, a utility function is employed to determine which of the incoming data samples should be labelled.
The most useful samples are queried as far as the labeling budget, which is defined as the percentage of the data samples that can be labeled per time unit, allows for it \cite{liobait2014ActiveLW}. 

Most existing \acrshort{al} strategies assume that labels are immediately available after data samples have been queried  \cite{Mohamad2016ActiveLF,liobait2014ActiveLW}. However, this is unrealistic  if labels cannot be calculated automatically and human experts need to be involved. 
As an example, in a typical factory setting, data samples for which labels are required within an   \acrshort{al} strategy are collected over some time period. Then an expert  inspects the current batch of data samples in infrequent intervals and provides some or all labels.
Hence a queried  label is available after some time period only, the so-called \textit{verification latency} \cite{Marrs2010TheIO}.The verification latency varies from sample to sample and it is unknown a priori. 
The presence of verification latency can be particularly problematic for machine learning approaches which deal with streaming data subject to concept drift, because a label might already be wrong when it arrives.
To the best of our knowledge, the effect of unknown verification latency on \acrshort{al} strategies is currently widely unexplored  \cite{Pham2021StreambasedAL}.
Further, most \acrshort{al} strategies assume a homogeneous budget over time, albeit  label requests might be particularly beneficial right after a detected drift event  \cite{Krawczyk2018CombiningAL,liobait2014ActiveLW}.


In this work, we explore the influence of verification latency in the presence of concept drift in \acrshort{al}.
We investigate limitations of existing \acrshort{al} strategies to estimate the utility of the current sample under a priori unknown  verification latency.
We focus on two aspects to improve this performance:
(1) A novel utility estimator called \acrfull{pr}, which concentrates on the feature space to infer the still unknown labels of queried samples, which is model-agnostic and latency independent.  (2) A dynamic budget allocation scheme, which distributes the overall labelling budget inhomogeneously over time as soon as driven a drift is detected. 
In the following, we investigate the performance of unsupervised and semi-supervised drift detectors under the effect of verification latency.
We test the performance by a thorough evaluation of several synthetic and real-world non-stationary datasets, with several realistic settings of verification delay, and we compare the proposed strategies against the \acrfull{sota} \cite{Pham2021StreambasedAL}.



\section{Related work} 
\textit{Verification latency} was introduced in \cite{Marrs2010TheIO} considering three different variants: \textit{null latency} refers to  a label for a selected sample arriving instantaneously --  this is the most common setting for \acrshort{al} in the literature; \textit{extreme latency}, i.e.\ the label is never available --  this setting is getting a lot of attention recently \cite{Umer2020ComparativeAO}; and \textit{intermediate latency}, with a finite delay between sample selection and label arrival -- this is common in many real-world applications, but has received only little attention in the literature \cite{Pham2021StreambasedAL}.
In \cite{Gomes2019MachineLF}, the authors list  research questions and approaches addressing verification latency  in  streaming data, but do not consider \acrshort{al} strategies.
\acrshort{al} strategies are considered in \cite{Kuncheva2008NearestNC,Parreira2021NaiveIW,Plasse2016HandlingDL}, where delayed labels are directly incorporated to the training loop of various models.
A utility estimation method similar to the one proposed here is considered in \cite{Pham2021StreambasedAL} which addresses the effect of delayed labels.
Yet the work  makes unrealistic assumptions about the label delay, e.g., fixed delay which is also known a priori;  the authors point out the necessity for more research on generalizations thereof \cite{Pham2021StreambasedAL,Serrao2018ActiveSL}.



Drift detection in presence of verification latency is getting some attention in recent years \cite{fahy2022scarcity,Mohamad2016ActiveLF}. 
Semi-supervised drift detection methods are rather prominent, which monitor the performance of the queried labeled in a specific task \cite{Krawczyk2018CombiningAL,liobait2014ActiveLW}.
But, with large verification delay recent labels might be lacking, leading to degraded performance of semi-supervised detection methods.
Unsupervised drift detectors within a  \acrshort{al} strategy is introduced in \cite{liobait2010ChangeWD}.
An adaptive labelling budget, where labeling ratio increases after a drift, was studied in \cite{Krawczyk2018CombiningAL}.
However, \textit{null latency} is assumed and semi-supervised drift detectors are used, which may not work in case of finite verification latency.


\section{Proposed Active Learning framework}

\begin{figure}[tb]
    \centering
    \includegraphics[width=0.75\linewidth]{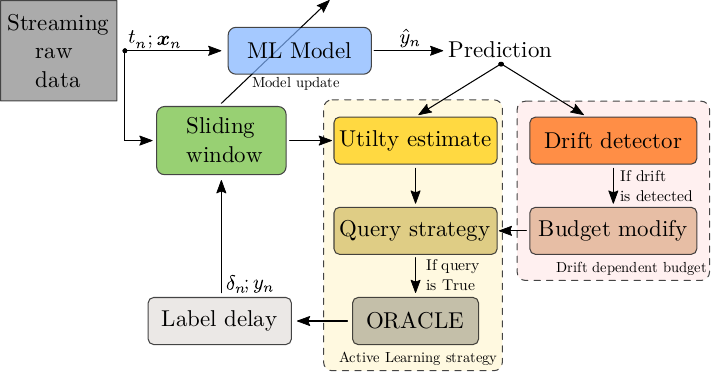}
    \caption{Proposed stream-based \acrfull{al} framework.}
    \label{fig:AL_framework}
\end{figure}

The processing flow for the proposed \acrshort{al} framework is shown in \fig{fig:AL_framework}.
Let $\mathcal{D}$ be a data stream, $\mathcal{D} = \{ (t_n, \bm{x}_n)\,|\, n \in \mathbb{N}  \} $, where each data sample is a \textit{d}-dimensional feature vector $\bm{x}_n$, which arrives at time $t_n$. 
We consider a classification task where label $y_n$ for the data sample can be actively obtained by querying an oracle.
The learning model is blindly updated at each time step $t_n$ using data (samples and available labels) from the sliding window  of the latest $l\in\mathbb{N}$ time steps $\mathcal{T}_n=[t_{n-l},t_{n}]$.
First, the utility of the current sample $\bm{x}_n$ is quantified based on the available information contained in the sliding window $\mathcal{T}_n$.
Then, the query strategy determines whether the data sample is queried or not, based on its utility and the budget \cite{liobait2010ChangeWD}.  The currently available budget $B(t)$, with $0\leq B(t_n)\leq 1$, is expressed as the probability that a data sample could be selected for querying at all.
The output of the querying strategy is realized as a Boolean variable $a_n$.
If the label is queried, $a_n=\m{True}$, then $\bm{x}_n$ is given to the oracle, so its label will be provided at the time $t_n+\delta_n$ and the sliding window $\mathcal{T}_n$ will be updated with the label information.
The delay $\delta_{n}$ is called \textit{verification latency}. 
In parallel to the utility calculation, a drift detector is employed. In case concept drift is detected, the budget for the query strategy is first increased substantially for a certain period of time, to adapt faster to the new  data statistics. After a while, when the model has hopefully adapted to the new data distribution, the budget is decreased to not exceed the number of total labeled samples, and later it returns to its nominal value. 


\subsection{Proposed utility estimator: PRopagate labels}
\begin{figure}[tb]
    \centering
    \begin{minipage}[t]{0.49\textwidth}
        \centering
        \includegraphics[width=1.\linewidth]{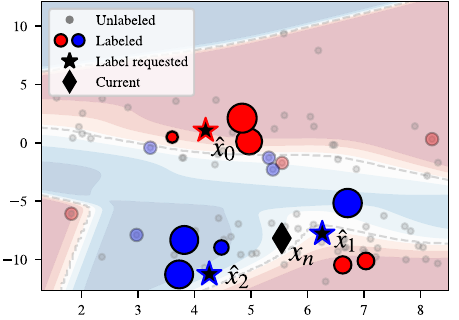}
        \caption{Schematic classifier decision regions.  Colored circles represent labelled data samples. The background color is the model confidence for each class. Circle size indicates weight $w_{i,j}$ where larger implies more recent.}
        \label{fig:PR_example}
    \end{minipage}\hfill
    \begin{minipage}[t]{0.49\textwidth}
        \centering
        \includegraphics[width=1.0\linewidth]{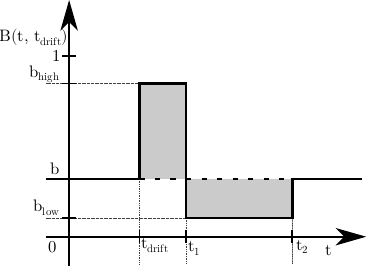}
        \caption{Drift detection driven budget distribution $B(t, t_{drift})$.}
        \label{fig:uneven_budget}
    \end{minipage}
\end{figure}

In a classical \acrshort{al} strategy, labels that have been queried, but are not yet available due to the verification latency $\delta_n$, would be ignored. As the utility function determining which samples to query is also not updated during that time,  this leads to an oversampling of high utility regions for querying.
A schematic example is shown in \fig{fig:PR_example} where the utility (taken as classification confidence) already led to submitting samples $\hat{x}_1$ and $\hat{x}_2$ to the oracle. Without labels, the current sample $\bm{x}_n$ would also have high utility and be a likely candidate for requesting its label. However, as soon as a label arrives for $\hat{x}_1$ or $\hat{x}_2$, not much information can  be gained by obtaining the label of $\bm{x}_n$.  

As a solution to this problem, we propose to PRopagate (PR) the spatial information of the queried labels to the neighboring unlabeled queried data samples, an idea inspired by \cite{Pham2021StreambasedAL}.
We estimate a still missing label with a weighted majority vote of the label of its $k$-Nearest Neighbor labels, restricted to samples from the sliding window $\mathcal{T}_n$.
The weight for each nearest neighbor depends on the arrival time of the labels  via
$w_{i,j} = \exp{(- \lambda (t_i - t_j)^2)}$, where $t_i$ and $t_j$ are respectively the timestamp of the queried sample and its neighbor $j$, with $j = 1 \dots k$. Therefore, weights for  newer labels are larger than older labels, which reflects that newer labels are less likely to be outdated as compared to older ones.  
Finally, the decision boundaries of the classifier are  updated using the true and predicted labels, and the utility of the current sample $\bm{x}_n$ is calculated.


In \fig{fig:PR_example} is shown an example of the proposed method (with $k=3$), the samples with requested but not yet arrived label have their neighbors highlighted. 
The marker size of their neighbors is proportional to $w$.
The samples $\hat{x}_0$ and $\hat{x}_2$ are assigned respectively to classes red and blue, since all their labeled neighbors belong to those classes.
The sample $\hat{x}_1$, even if the majority of its neighbors belong to the red class, is assigned to the blue class, due to of the weight strategy.

\subsection{Proposed budget strategy: dynamic budget allocation}
In order to increase the performance for non-stationary data streams   
we include  feedback between the \acrshort{al} strategy and the concept drift detection module.
We propose to use a concept drift-dependent distribution of the labeling budget over time, without exceeding the global budget limit for the complete data stream, i.e.\ the total number of queried labels.
We employ a piece-wise constant budget function  $B(t, t_{drift})$,
\begin{equation}\label{eq:B}
    B(t, t_{drift}) = 
    \begin{cases*}
      b_{\m{high}} & if $t_{drift} < t < t_1$ \\
      b_{\m{low}}  & if $t_1 < t < t_2$ \\
      b & otherwise
    \end{cases*}\:,
\end{equation}
which is sketched in \fig{fig:uneven_budget}. 
After  a detected drift, the budget increases to $b_\m{high}$ for a time period $t_1 - t_{drift}$, and then decreases to $b_\m{low}$ for a time period $t_2 - t_1$. After the time $\Delta T=t_2-t_{drift}$ the budget is back to $b$, which is the value for no detected drift. 
In order to have the same total number of queries to the oracle with and without drift, not all parameters of this budget can be chosen  freely, but the condition $\int_{t_{drift}}^{t_{drift}+\Delta T} B(t) \, dt \overset{!}{=} b \Delta T$ needs to be fulfilled. Given am overall budget $b$, we freely adjust parameters $b_\m{high}$,  $b_\m{low}$ and $\Delta T$, which leads to the time points $t_1$ and $t_2$:
\begin{equation}\label{eq:deltas}
        t_1 = t_{drift}+\Delta T \frac{b - b_\m{low}}{b_\m{high} - b_\m{low}} \,,\quad 
        t_2 = t_{1}+ \Delta T \frac{b_\m{high} - b}{b_\m{high} - b_\m{low}}
\end{equation}
In this way, the total rate of labels queried by the \acrshort{al} strategy is the same as with constant budget $b$.
Yet it is inhomogeneously distributed, i.e., after a drift, the labelling budget is substantially increased. This leads to a quicker model update but needs to be compensated by reducing the budget later on to arrive at the same average label rate. 

\begin{algorithm}[t]
\DontPrintSemicolon
\footnotesize
\KwInput{Classifier $f$, drift detector $\Psi$, utility function $UL$, query strategy $QS$, budget function $B$ , budget parameters $b_\m{high}$, $b_\m{low}$, $\Delta T$ }

$f$, $\Psi$, $UL$, $QS \gets$ initialize \;
$t_1$, $t_2 \gets $ Eq.~\eqref{eq:deltas} \;
$t_{drift} \gets +\infty $ \;

\For{$n$ in ${1, \dots }$} {
$(t_n, \bm{x}_n) \gets$ retrieve new data sample from data stream\;
$\hat{y}_n = f(\bm{x}_n)$ \tcp*{Get label prediction}

\uIf{\textit{$\Psi(\mathcal{T}_n) = \m{True}$}}{
$t_{drift} = t_n $ \tcp*{Drift detected}
}

$b_n \gets B(t_n, t_{drift}) $ \tcp*{Eq.~\eqref{eq:B}: Drift dependent budget modification}

$u_n \gets UL(\bm{x}_n, \hat{y}_n, \mathcal{T}_n, f)$ \tcp*{Get utility for current sample}
$a_n \gets QS(b_b, u_n)$ \tcp*{Query sample}
\uIf{$a_n = True$}{
ask oracle for label of $\bm{x}_{n'}$ (label $y_n$ will be provided at $t_n + \delta_n$)}

$\mathcal{T} \gets (t_n, \bm{x}_n, y_n)$ \tcp*{Update sliding training window}
$f, \Psi \gets $ update with $\mathcal{T}$ \;

}
\caption{Proposed stream-based AL framework}\label{alg:AL}
\end{algorithm}

\section{Experimental setup}

We perform all the experiments with the proposed stream-based AL framework sketched in Alg.~\ref{alg:AL}.
We evaluate the proposed method on synthetic and real-world datasets, listed in Table~\ref{tab:data}.
For benchmarking purposes, concept drift is artificially introduced and controlled after 50\% of the data stream, by corrupting the most informative features in a gradual manner, as described in \cite{Castellani2021TaskSensitiveCD}.

We evaluate the model with the prequential evaluation \cite{Gama2009IssuesIE} (test-then-train) over all instances of the data stream.
We use the Parzen Window Classifier (PWC). In order to make a fair comparison with the existing literature, we follow the benchmark framework of \cite{Pham2021StreambasedAL}.
The classifier is first initialized with the first 100 samples of each data stream, then it is trained used a sliding window of $l=500$ samples.
As stream-based querying strategies, we use \textit{Split} \cite{liobait2014ActiveLW} and \textit{Probabilistic Active Learning (PAL)} \cite{Kottke2015ProbabilisticAL}.
We also use the baseline strategy \textit{Random Selection (rand)}, that randomly samples instances according to the given budget.
We compare the proposed sample utility estimation strategy (PR) to the current \acrshort{sota} introduced in \cite{Pham2021StreambasedAL}, which is referred to as  `\textit{forgetting and simulating incoming labels with bagging (FS.B)}'.

We investigate the influence of variable verification latency by sampling the actual delay $\delta_n$ for each label query from a predefined distribution. We tested two representative distributions,  a uniform distribution with  $\delta_n\sim \mathcal{U}(50, 50+\delta)$, and a truncated normal distribution,  $\delta_n\sim \max(0,\mathcal{N}(\delta, 50))$ (truncation refers to the fact that we set a negative delay to zero).
We tested various parameter values for $\delta=[0, 50, 100, 150, 200, 300]$, and several budgets levels $b$ of 5\%, 10\%, 15\%, 20\%, 25\%, 40\%, 50\%, 75\% and 100\% as an upper bound of performance.
Unless explicitly reported, we set the number of neighbors for the label propagation to $k=3$ and the weighting coefficient of $\lambda=0.01$. 
The budget modification time is set to $\Delta T = 1000$ and modified budgets are $b_\m{high}=4b$ and $b_\m{low}=b/2$. 
We use two popular semi-supervised drift detection algorithms, {Drift Detection Model} (DDM) \cite{Gama2004LearningWD} and {Adaptive Windowing} (ADWIN) \cite{Bifet2007LearningFT}. These act only using the labeled samples coming from the active learning \cite{Krawczyk2018CombiningAL}. We also employ the unsupervised {Hellinger Distance Drift Detection Model} (HDDDM) drift detection method \cite{Ditzler2011HellingerDB}, operating directly on data samples in the feature space of the input data.
For each algorithm and dataset, we monitor the accuracy of the whole data stream.
All experiments have been repeated 50 times with different random seeds. 
The source code used in the experiments is publicly available on \verb|http://github.com/Castel44/AL_delay|.

\begin{table}[tb]
    \centering
    \footnotesize
    \caption{Details of the data streams used for training.}
    \begin{tabular*}{\linewidth}{l @{\extracolsep{\fill}} l l l l l l}
    \toprule
        \textbf{Dataset} & \textbf{Instances} & \textbf{Features} & \textbf{Classes} &\textbf{Data type} &\textbf{Drift type}\\
        \midrule
        \textit{RBF\_2\_2} \cite{Pham2021StreambasedAL} & 4000 & 2 & 2 & \multirow{4}{*}{Synthetic} & \multirow{4}{*}{Induced}\\
        \textit{RBF\_10\_4} \cite{Pham2021StreambasedAL} & 4000 & 10 & 4 \\
        \textit{hyperplane} \cite{Gama2004LearningWD} & 4000 & 2 & 2 \\
        \textit{stagger} \cite{Gama2004LearningWD} & 4000 & 2 & 2 \\
       \midrule
        \textit{wine} \cite{Castellani2021TaskSensitiveCD} & 6497 & 12 & 2 & \multirow{3}{*}{Real-world} & \multirow{3}{*}{Induced}\\
        \textit{digits08} \cite{Castellani2021TaskSensitiveCD} & 1499 & 16 & 2 \\
        \textit{digits17} \cite{Castellani2021TaskSensitiveCD} & 1457 & 16 & 2\\
        \midrule
        \textit{Luxembourg} \cite{liobait2011CombiningSI}& 1901 & 31 & 2 & \multirow{2}{*}{Real-world} & \multirow{2}{*}{Unknown}\\
        \textit{Weather} \cite{Ditzler2013IncrementalLO} & 18159 & 8 & 2\\
    \bottomrule
    \end{tabular*}
        \label{tab:data}
\end{table}

\section{Results and discussion}
Due to space restrictions, we report only a  representative subset of all results, which clearly  demonstrate the qualitative trends obtained in all the experiments.

\subsubsection{Effect of unknown label delay in existing AL strategies.}
\begin{figure}[tb]
    \centering
    \begin{minipage}[t]{0.49\textwidth}
        \centering
        \includegraphics[width=0.99\linewidth]{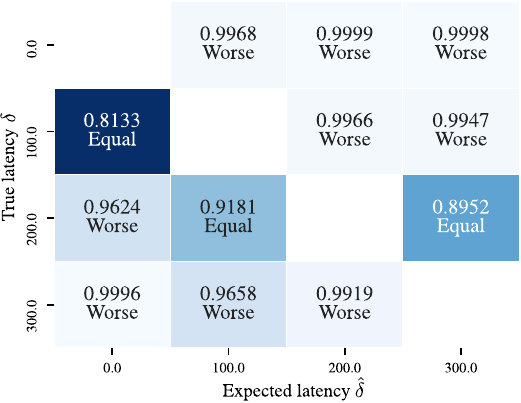}
        \caption{P-values for \textit{FS.B+pal} with true latency vs expected latency.}
        \label{fig:pval}
    \end{minipage}\hfill
    \begin{minipage}[t]{0.49\textwidth}
        \centering
        \includegraphics[width=0.90\linewidth]{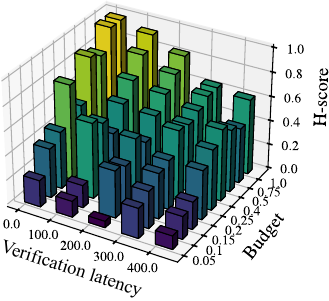}
        \caption{Semi-supervised drift detector (ADWIN) performance.}
        \label{fig:drift_detector_3d}
    \end{minipage}
\end{figure}

The \acrshort{sota} utility estimator for \acrshort{al} strategies requires knowledge of the verification latency in advance.
We analyze the performance of the \textit{FS.B+pal} strategy \cite{Pham2021StreambasedAL}, under the influence of verification latency sampled from the truncated normal distribution, $\delta_n\sim\max(0, \mathcal{N}(\delta, 50))$, where the parameter value $\delta$ is unknown in advance. 
In the experiments, when the expected latency  $\hat{\delta}$ of \textit{FS.B+pal} is not equal to the true latency,  $\delta \neq \hat{\delta}$, we witness an average accuracy drop of 0.87\%, with 95\% 
confidence interval (0.21\% - 2.15\%).
For each value of true latency $\delta$, we compare the performance obtained with the correct match of the true latency against expected latency values. 
In \fig{fig:pval}, we report p-values of the non-parametric statistical \textit{Mann-Whitney U-Test} \cite{mcknight2010mann} with $\alpha$ level of 0.05 for the dataset \textit{RBF\_4\_10} based on the null hypothesis that the performance is not affected by differing latencies.
It can be clearly observed that there are statistical differences when the actual and the expected latency differ, and the null hypothesis is rejected for most of the combinations.

\subsubsection{Experimental results with proposed utility estimator PR.}

\begin{figure}[tb]
    \centering
    \includegraphics[width=1.\textwidth]{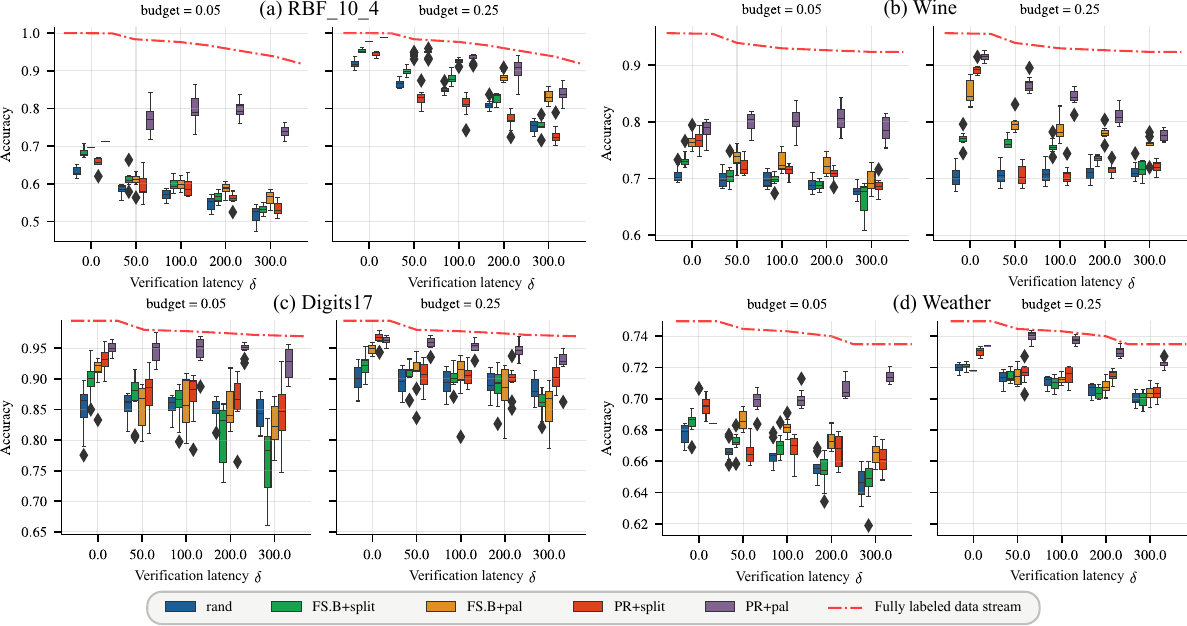}
    \caption{Results with fixed budget distribution.}
    \label{fig:static_budget}
\end{figure}

We compare the performance achieved with the proposed \acrshort{pr}, against the \acrshort{sota} utility estimator FS.B \cite{Pham2021StreambasedAL}. 
We combine it with two popular query strategies, \textit{split} \cite{liobait2014ActiveLW} and \textit{pal} \cite{Kottke2015ProbabilisticAL}. 
\fig{fig:static_budget} shows the results with low (5\%) and medium (25\%) budget for different levels of varying verification latency $\delta_n\sim \max(0,\mathcal{N}(\delta, 50))$.
Note that in this case even with $\delta=0$, about half of the labels  arrive with some delay.
To compare \acrshort{pr} to the best possible \acrshort{sota} algorithm, we use the real expected latency $\hat\delta=\delta$ for the latter.
As upper limit on the performance, the dashed red line shows the results when every data sample is labeled, i.e.\ 100\% budget. 
As expected, the classification accuracy decreases with the increase of the verification latency.
The proposed utility estimator consistently outperforms the \acrshort{sota} for the same querying strategies.
The best performance is achieved combining the \acrshort{pr} utility estimator with the \textit{pal} \acrshort{al} strategy.
Also, for $\delta=0$ the \acrshort{pr} slightly outperform the \acrshort{sota}.
Because the \acrshort{sota} assumes zero delay, leading to reduced performance, while the PR utility estimator makes no assumption on the expected delay.
This confirms the expectation that using the current classification model to estimate the latest queried labels, as done in the \textit{FS.B} query estimation \cite{Pham2021StreambasedAL}, leads to reduced performance in case of drift, while relying on the recently sampled labels utilizes the available information more efficiently.  
It is interesting to note that already for 25\% budget the proposed method comes close to the best cases with 100\% budget for the three datasets \texttt{RBF\_10\_4}, \texttt{Digits17} and \texttt{Weather}.


\subsubsection{Drift detection performance.}
In \acrshort{al} strategies, semi-supervised drift detectors can be used to monitor the error rate of the requested labeled samples \cite{Krawczyk2018CombiningAL,liobait2014ActiveLW}.
We empirically show that under the influence of verification delay, and low budget, such drift detectors have low performance. Either the drift is not detected, or it is detected too late, when the model already recovered due to the \acrshort{al} process.
We use the H-score \cite{Castellani2021TaskSensitiveCD} as drift detection metric, as it takes in account the real detection of a change event and its detection delay. 
In \fig{fig:drift_detector_3d} we report the ADWIN detection performance on the dataset \texttt{RBF\_10\_4}.
A good level of performance ($H > 0.75$) is only achieved with budget above 0.25, and with latency less than 200.
Hence, hereafter, we use the unsupervised drift detector (HDDDM) which is independent of budget or latency, and it is able to promptly detect the drifts in the distribution of the feature space.

\subsubsection{Drift detection driven budget distribution.}

\begin{figure}[tb]
    \centering
    \includegraphics[width=1.\linewidth]{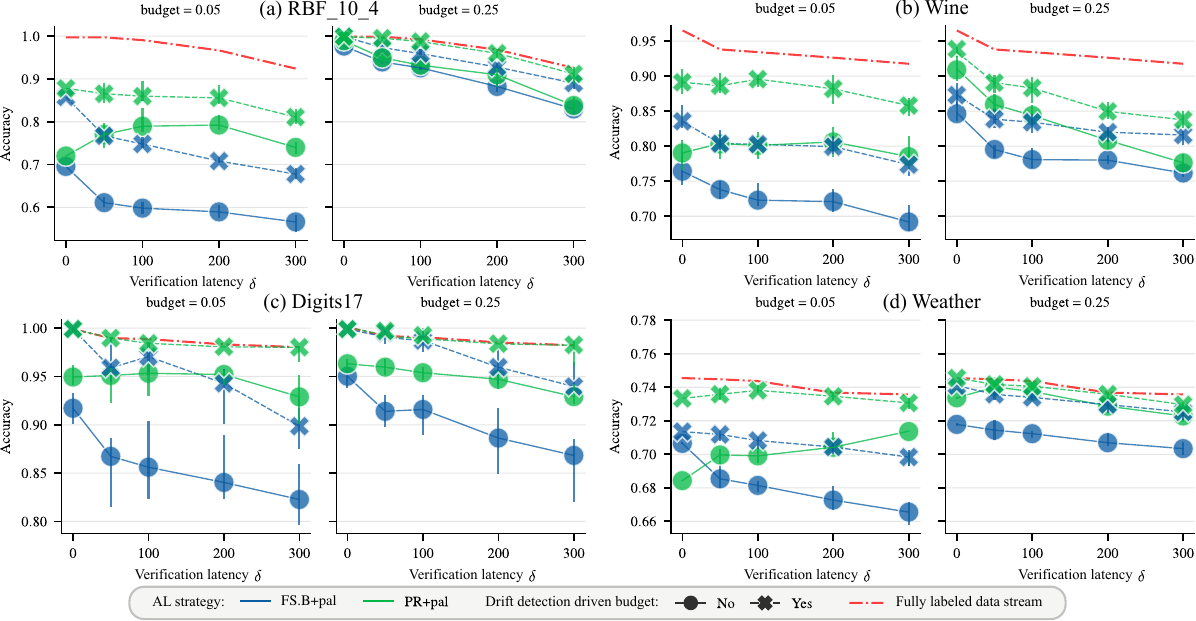}
    \caption{Results of the proposed drift detection driven budget distribution.}
    \label{fig:detector}
\end{figure}

In \fig{fig:detector}, we report the performance of the proposed \acrshort{al} framework with dynamic budget allocation after a detected drift, as described in Eq~\eqref{eq:B}. 
We compare it to using a fixed static  budget $b$, where the total number of queried labeled sample is the same in both approaches.
Again, the red dashed line is the upper bound of performance, obtained with a fully labeled data stream. 
For readability, we only report the best performing query strategy (\textit{pal}) of \fig{fig:static_budget}, and we compare the proposed PR with \textit{FS.B} utility estimator.
It is clear that the dynamic budget allocation, after a detected drift, increases the performance for both approaches considerably (crosses compared to full circles). 
Also, the proposed method for label propagation consistently outperforms the \acrshort{sota}.
Remarkable, the proposed budget strategy is able to obtain accuracy level comparable to using a 100\% labeled data stream, with as little as 5\% of labeled data.
This demonstrates that by carefully labeling  instances, we are able to achieve competitive model accuracy with much lower cost.

\comment{
\begin{figure}[tb]
    \centering
    \includegraphics[width=0.75\linewidth]{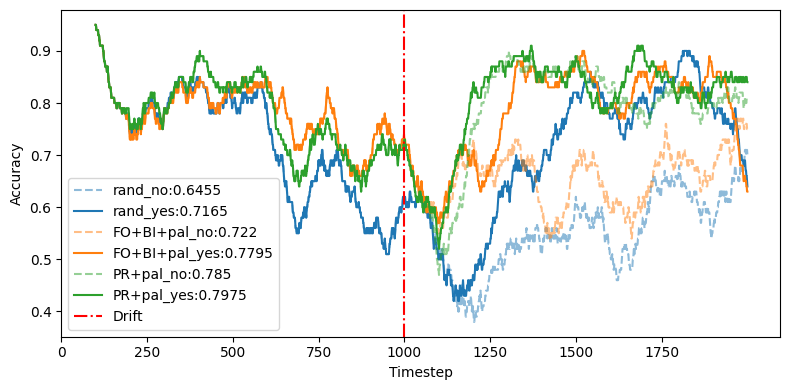}
    \caption{Accuracy curve. \andrea{Remove?}}
    \label{fig:accuracy_curve}
\end{figure}
}

\begin{figure}[tb]
    \centering
    \begin{minipage}[t]{0.54\textwidth}
        \centering
        \includegraphics[width=1.\linewidth]{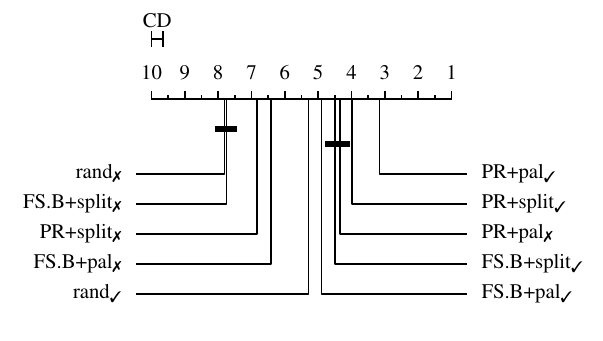}
        \caption{Critical distance diagram summarizing all experiments.}
        \label{fig:cd}
    \end{minipage}\hfill
    \begin{minipage}[t]{0.44\textwidth}
        \centering
        \includegraphics[width=0.77\linewidth]{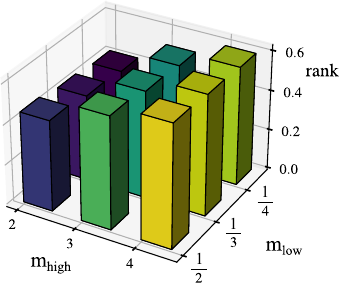}
        \caption{Average performance rank of modified budget hyper-parameters.}
        \label{fig:rank_mn}
    \end{minipage}
\end{figure}

As a summary, in order to assess the statistical significance of the results, we use the Friedman non-parameteric test with 0.05 confidence level, followed by Nemenyi post-hoc test. 
We report in \fig{fig:cd} the critical differences plot \cite{demsarCD06} of our experiments, accumulated over all the datasets, budget and latency levels.
The querying strategies combined with the proposed utility estimator (PR) are significantly better than the \acrshort{sota} counterpart. 
The \acrshort{al} strategies using the proposed drift driven budget distribution (subscript \cmark) are in all cases significantly better than the corresponding methods with the static budget allocation (subscript \xmark).

\subsubsection{Ablation studies.}
We performed ablation studies on the hyper-parameters for the modified budget of Eq.~\eqref{eq:B}  and  used $b_\m{high}=b\,m_\m{high} $ and $b_\m{low}=b \, m_\m{low}  $ with $m_\m{high}\in[2,3,4]$ and $m_\m{low}\in[\tfrac12,\tfrac13,\tfrac14]$.
The results of the rank of the proposed approach averaged over all datasets and delays are shown in \fig{fig:rank_mn}, where  we used a window $\Delta T = 1000$ timesteps.  Therefore, the best performing setting of $m_\m{high}=4$ and $m_\m{low}=\tfrac12$ has increased budget for $t_1 - t_\m{drift} =143$, and  decreased budget for $t_2 - t_1=857$ timesteps.

\section{Conclusion and future work}
In this work, we addressed the problem of \acrfull{al} under finite, variable, and unknown verification latency.
We proposed \acrfull{pr}, a model-agnostic utility estimator strategy for \acrshort{al}, which uses the known labels to infer labels for queried but not yet arrived labels.
The utility for querying subsequent labels is then calculated with all, known and estimated, labels.
We also propose to use a dynamic allocation of the labeling budget over time, driven by the detection of concept drift events. After a drift detection, we increase the budget, then we decrease it in order to meet the total budget labeled samples.

Experimental results with real-world data streams and realistic settings of latency, showed that existing \acrshort{al} strategies are sub-optimal when the amount of latency is unknown.
By using the proposed \acrshort{pr} we consistently outperform the \acrshort{sota}.
We also empirically proved that under the effect of verification latency, semi-supervised concept drifts detectors have poor performance.
Then, we proved that the proposed drift detection driven budget allocation improves the performance of the \acrshort{al} strategies. 
With the proposed method, we showed that is possible to achieve similar results as we use the fully labeled data stream, with as little as 5\% of labeled samples.
We thoroughly analyzed the dependency of the introduced hyperparameters and identified a range of values for robust and good operation.

Even though the proposed methods are model-agnostic, for this article, we only used the Parzen Window Classifier, and we used the classification confidence as utility measure.
In future, we extend to apply the proposed method to semi-supervised Deep Learning classifiers and to use other utility measures e.g. information gain. 
As shown, the dynamic budget strategy works well in the current setting, where only  one abrupt drift occurs. An open question for future work is the investigation of the proposed approach in  situations where multiply drifts might occur during the dynamic budget adjustment period, or where a continuous drift occurs over a longer period of time. 

\bibliographystyle{splncs04}
\bibliography{biblio.bib}

\end{document}